\documentclass[nonacm,sigconf]{acmart}

\AtBeginDocument{%
  }

\renewcommand\footnotetextcopyrightpermission[1]{}

\usepackage{amsmath}
\usepackage{algorithm}
\usepackage{algpseudocode}
\usepackage{amsfonts}

\usepackage{amsmath}   
\usepackage{amsfonts}  

\usepackage{pgfplots}
\pgfplotsset{compat=1.18}
\usepackage{caption}
\usepackage{subcaption}

\usepackage{enumitem}
\setlist{nosep}

\begin{document}

\title{Designing for the Next Click:  Bandits for Real-Time Page Layout }





\author{Bhavtosh Rath}
\affiliation{%
  \institution{Target Corporation}
  \city{Minneapolis}
  \state{MN}
  \country{USA}
}
\email{bhavtosh.rath@target.com}

\author{Harshith Narasimhamurthy}
\affiliation{%
  \institution{Target Corporation}
  \city{Minneapolis}
  \state{MN}
  \country{USA}
}
\email{harshith.narasimhamurthy@target.com}

\author{Bob Eisinger}
\affiliation{%
  \institution{Target Corporation}
  \city{Minneapolis}
  \state{MN}
  \country{USA}
}
\email{bob.eisinger@target.com}

\author{Cole Stiegler}
\affiliation{%
  \institution{Target Corporation}
  \city{Minneapolis}
  \state{MN}
  \country{USA}
}
\email{cole.stiegler@target.com}

\author{Adnan Awow}
\affiliation{%
  \institution{Target Corporation}
  \city{Minneapolis}
  \state{MN}
  \country{USA}
}
\email{adnan.awow@target.com}

\author{Amit Pande}
\affiliation{%
  \institution{Target Corporation}
  \city{Minneapolis}
  \state{MN}
  \country{USA}
}
\email{amit.pande@target.com}



\renewcommand{\shortauthors}{Bhavtosh Rath et al.}

\begin{abstract}
  E-commerce platforms increasingly personalize user experiences through machine learning, yet page layout decisions remain dominated by static rules and manual curation. We present a scalable bandit-based system that optimizes product page layouts in real time while preserving human control over design intent. A contextual bandit model  dynamically selects the most effective layout for each session using user, item, and category-level features. The system leverages a LinUCB-based policy to balance exploration and exploitation as it learns from live user interactions. The architecture is designed for seamless integration into large-scale web serving stacks, supporting low-latency inference and continuous model updates. The system was first tested on entry product pages. In online A/B deployments on a major retail platform, our approach achieved positive lifts in  session-level performance metrics over a strong heuristic baseline. Our results demonstrate that contextual bandits can effectively optimize visual and structural aspects of product discovery for user engagement, providing a scalable path toward learning-to-design the web.
\end{abstract}





\maketitle

\section{Introduction}
The design of e-commerce page layouts significantly impacts user engagement and conversion rates, yet optimizing these layouts remains a challenging problem. Traditional A/B testing approaches require manual hypothesis generation, fixed experiment durations, and often fail to account for changing user preferences over time. Meanwhile, the explosion of product catalogs, page layouts, and diverse user segments creates a combinatorially large space of possible layout configurations that is infeasible to explore exhaustively.
We present a framework for e-commerce page layout optimization that leverages multi-armed bandit algorithms to continuously learn and adapt layout configurations in real-time. 

Our key contributions are:
\begin{enumerate}
\item Contextual Bandit Formulation: We formalize the page layout optimization problem as a linear-contextual bandit, where context includes user, item, and category features, enabling personalized layout selection.
\item Production-Scale Online Deployment: We introduce a hierarchical representation of page layouts as compositions of modules (hero sections, product grids, recommendation carousels), allowing efficient exploration of the combinatorially large layout space.
\item Real-World Deployment: We demonstrate the effectiveness of our approach through a large-scale deployment on an e-commerce retailer serving millions of users daily, achieving a 4.9\% improvement in click-through-rate and a 1.5\% lift in cart adds to curated layouts. 
\end{enumerate}
The evaluation was conducted on the company's external traffic product detail pages—a critical entry point for users arriving via search engines, social media, and other external channels rather than through the site’s internal navigation. These pages present a uniquely challenging environment: they historically exhibit higher bounce rates compared to internal traffic pages, and user context is severely limited since most visitors are not logged in, leaving us without access to their browsing history or the original search queries that brought them to the platform.

Currently, the recommendation carousels on these pages use a static configuration that remains identical across all product categories and user segments. This one-size-fits-all approach represents a significant optimization opportunity. By deploying a contextual bandit algorithm, we can transform these static carousels into a dynamic, self-learning system. 
The positive lift observed in this challenging environment is particularly encouraging. Having demonstrated effectiveness where user context is most limited, we plan to expand the system to other digital touchpoints where richer contextual signals (logged-in status, session history, personalization data) should enable even stronger performance gains.

\begin{figure*}[ht]
\centering
\includegraphics[width=0.95\linewidth]{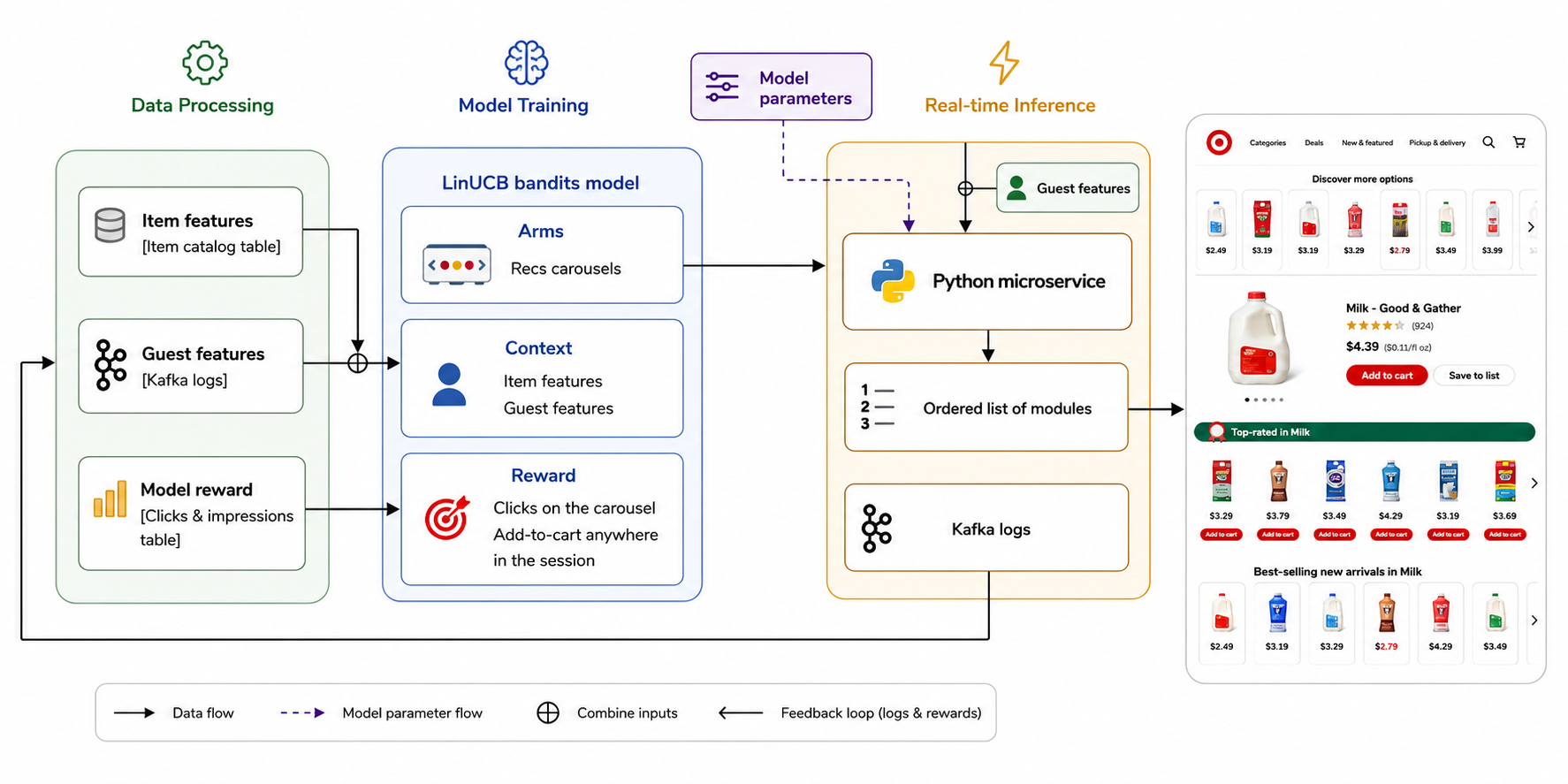}

\caption{Architecture of Page Layout Optimization. Item metadata \& user behavior logs provide context, while engagement data defines reward. LinUCB model ranks page modules in real time, generating layouts optimized for user engagement.}
\label{fig:architecture}
\end{figure*}



\section{Related Work}

Personalization has been extensively studied through contextual multi-armed bandit (CMAB) algorithms, which balance exploration and exploitation to optimize user engagement. Early work extended contextual bandits beyond e-commerce; for example, Lan et al.~\cite{lan2016contextual} applied them to adaptive education systems, establishing CMABs as an efficient paradigm for real-time personalization.

Among foundational contributions, Li et al.~\cite{li2010contextual} introduced LinUCB for news recommendation, demonstrating effective personalization using user and item features. Subsequent algorithms such as Epoch-Greedy~\cite{langford2008epoch} and the scalable frameworks of Agarwal et al.~\cite{agarwal2014taming} improved computational efficiency for large-scale deployments. Comprehensive surveys~\cite{bouneffouf2019survey} highlight the versatility of bandits across domains including recommendation, information retrieval, and healthcare.

Agarwal et al.~\cite{agarwal2016making} deployed a decision service to operationalize with low technical debt, while Foster et al.~\cite{foster2018practical} proposed regression-based contextual bandits for production-scale systems.

Beyond linear models, neural contextual bandits have emerged to capture nonlinear user–item relationships. Zhou et al.~\cite{zhou2020neural} introduced Neural UCB, integrating deep learning with confidence-based exploration, and Kassraie and Krause~\cite{kassraie2022neural} proposed NN-UCB with sublinear regret guarantees via neural tangent kernels. Building on these ideas, Shi et al.~\cite{shi2023deep} combined deep features with LinUCB (DeepLinUCB), and Yi et al.~\cite{yi2023online} presented Online Matching, a large-scale bandit system unifying offline pretraining and real-time learning at YouTube.

Recent work has also considered recommendation at the page level, where the objective extends beyond ranking items within a single recommendation list to selecting and ordering multiple recommendation modules. Lo et al.~\cite{lo2021page} presented a production system for page-level optimization of e-commerce recommendations, dynamically selecting and ordering recommendation modules on item-detail pages. More broadly, whole-page optimization has been formulated as a sequential decision problem; for example, prior work has explored reinforcement-learning approaches for jointly optimizing page layouts rather than independently ranking individual recommendation lists~\cite{qin2022automate}. These systems highlight the growing importance of optimizing the composition of the overall page experience rather than treating each recommendation module in isolation. 

Our work builds on this literature but focuses on a narrower, production-oriented setting. In our initial deployment, we use a contextual bandit to select and rank the top seven recommendation modules from a set of eligible modules on product detail pages. Our contribution is therefore not a new bandit algorithm or a general solution to whole-page optimization; rather, we study the practical deployment of contextual bandits for online module ranking, with an emphasis on exploration--exploitation, low-latency inference, continuous learning, and integration with a large-scale e-commerce serving stack. We view this deployment as a first step toward broader full-page layout optimization, where heterogeneous page components could eventually be jointly selected, ordered, and personalized.

\section{System}
\label{sec:linucb}

The webpage layout is decomposed into \textit{modules}—portable experience units that can be displayed, hidden, or reordered across pages. We propose a module-ranking task that seeks the optimal module order to maximize engagement per user. Offline supervised ranking models learn from historical data but cannot explore new orderings \cite{joachims2017unbiased}, while full reinforcement learning captures sequential dependencies but is complex and data-hungry \cite{liu2018deep}. Contextual bandits provide a practical middle ground: they leverage contextual signals to adapt rankings in real time through online feedback, efficiently balancing exploration and exploitation to directly optimize engagement.

In a contextual bandit, at each interaction \( t \), the agent observes a context vector \( x_t \in \mathbb{R}^d
 \) and must select an ordered slate of modules (arms) from an eligible list of modules \( \mathcal{A} \). Upon selection, the model receives a reward \( r_t \). The objective is to maximize cumulative reward over time by balancing exploration (trying new arms) and exploitation (using the best-known arms).
We employ the disjoint LinUCB contextual bandit model for this application. 
Compared to alternative approaches such as $\varepsilon$-greedy or Thompson Sampling, 
LinUCB provides an effective balance of \textit{simplicity}, \textit{fast inference}, \textit{rapid learning}  and \textit{interpretability}—all of which are critical for our real-time use case.

In the LinUCB algorithm, the expected reward for arm \( a \) is assumed to be a linear function of the context:
$\mathbb{E}[r_t | x_t, a] = x_t^\top \theta_a,$
where \( \theta_a \in \mathbb{R}^d \) represents the parameter vector associated with arm \( a \).  
Details of the LinUCB method are given in Algorithm~\ref{alg:linucb-slate}, where
$A_a$ and $b_a$ are the per–arm sufficient statistics for ridge regression.

\begin{algorithm}[tb]
\caption{Disjoint LinUCB-Slate}
\label{alg:linucb-slate}
\begin{algorithmic}[1]
\Require Slate size $L$, exploration parameter $\alpha > 0$
\State \textbf{Initialize:} For each arm $a$, set $A_a \gets  I_d$, $b_a \gets 0_d$
\For{$t = 1, 2, \ldots, T$}
    \State Receive eligible set $\mathcal{A}$ and contexts $\{x_{t,a} \in \mathbb{R}^d : a \in \mathcal{A}\}$
    \For{each $a \in \mathcal{A}$}
        \State $\hat{\theta}_a \gets A_a^{-1} b_a$ \Comment{use cached $A_a^{-1}$ if available}
        \State $p_{t,a} \gets x_{t,a}^\top \hat{\theta}_a + \alpha \sqrt{x_{t,a}^\top A_a^{-1} x_{t,a}}$
    \EndFor
    \State \textbf{Ranking:} Let $S_t$ be the $L$ arms with largest $p_{t,a}$ values; display them sorted in descending $p_{t,a}$ (top $\to$ bottom)
    \State Observe reward $r_{t}$ for displayed $a \in S_t$
    \For{each $a \in S_t$}
        \State $A_a \gets A_a + x_{t,a} x_{t,a}^\top$
        \State $b_a \gets b_a + r_{t} x_{t,a}$
    \EndFor
\EndFor
\end{algorithmic}
\end{algorithm}



To mitigate the bias due to position (i.e. modules on top of the page do not require scrolling effort) an optional inverse propensity weighting is applied to the reward:
$r'_t = r_t \cdot \min\left(1/{\max\left(\frac{\text{imp}_{d_t}}{\text{imp}_0}, \varepsilon\right)}, C_{\max}\right),$
where \( \text{imp}_{d_t} \) is the impression count at position \( d_t \), \( \text{imp}_0 \) is the reference (top) position, and \( C_{\max} \) is a clipping threshold.  

\label{sec:reward_logic}
In future applications, arms can be any of the modules present in the product page such as display product image, add to cart button, reviews etc. However, in our first experiment, we limited our modules to a set of product recommendation modules, 7 of which are shown to a user. We curated 18 possible modules and let the contextual bandit decide the rank order of 7 to be shown to the user. 
The reward signal tries to capture the interaction of the user with a provided module. If the module was clicked, we provide a reward of 1, whereas no reward (0) is recorded if there is no action. The context features used are summarized in Table~\ref{tab:context_features}. 

\begin{table}[tb]
\centering
\small
\caption{Context features used in the LinUCB model, metadata indicates datatype and data source.}
\hspace*{-0.02\textwidth} 
\begin{tabular}{@{}p{0.03\textwidth}p{0.37\textwidth}p{0.07\textwidth}@{}}
\toprule
\textbf{\scriptsize{}} & \textbf{Description} & \textbf{metadata} \\ \midrule
$f_{1}$--$f_{10}$ & \textit{Breadcrumb embeddings}: dense sentence embeddings of each item's hierarchical category path (e.g., \texttt{furniture $\rightarrow$ living room furniture $\rightarrow$ sofas \& couches}), reduced to 10 dimensions using PCA. These capture semantic similarity across related product families, enabling consistent placement behavior for conceptually similar categories. & Numeric, Item \\[4pt]
$f_{11}$ & \textit{Normalized regular price}: reflects relative affordability rather than absolute cost, allowing user engagement patterns to generalize across categories with different price scales. & Numeric, Item \\[4pt]
$f_{12}$ & Indicates if item is newly introduced. & Bool, Item \\[4pt]
$f_{13}$ & Indicates if item is trending in user views. & Bool, Item \\[4pt]
$f_{14}$ & Indicates if item is on some sale/clearance/deal & Bool, Item \\[4pt]
$f_{15}$--$f_{18}$ & One-hot encoded brand group vector (\texttt{company-Owned}, \texttt{National}, etc.), capturing brand-conditioned preferences  & Categorical, Item \\[4pt]
$f_{19}$--$f_{21}$ & Binary recency indicators denoting whether the specific product page was viewed, purchased, or added to cart within the past 14 days by same visitor.  & Bool, User \\[4pt]
$f_{22}$--$f_{24}$ & Category-level recency indicators capturing whether any item within the same category was viewed, purchased, or added to cart recently.  & Bool, User \\ 
\bottomrule
\end{tabular}
\label{tab:context_features}
\end{table}

\subsection{Training \& Inference}
The model is trained and updated every day on the past N day window.
A dedicated microservice orchestrates the inference workflow. This service reads item feature data and model parameter files from a cloud bucket and constructs context features in real time using user interaction data. Simultaneously, Kafka logs are streamed and persisted to disk to facilitate the extraction of user engagement features for continuous model improvement and retraining. Our system follows a microservice architecture in which the client-facing API communicates with the Contextual Bandit (CB) microservice using the gRPC protocol. The services are deployed on Kubernetes with an autoscaler to dynamically manage traffic spikes. The microservice logs are asynchronously logged to a Kafka queue and stored to disk. Figure \ref{fig:architecture} explains the overall workflow. The platform's frontend architecture relies heavily on Server-Side Rendering (SSR) to optimize first-pixel rendering and minimize bounce rates—delays of even a few milliseconds can significantly impact user engagement. This made real-time API latency our primary constraint when introducing the contextual bandit system. To minimize the performance impact of the dynamic API calls required by the contextual bandit system, we implemented several latency optimizations. Through microservice-level caching of item and category-specific features, combined with O(1) lookups of user features from Kafka topics and asynchronous event log writes, we achieved a p95 latency of under 25ms. Additionally, our choice of LinUCB as the core algorithm provides ultra-low inference latency through efficient linear computations, ensuring the benefits of dynamic personalization don't compromise page load performance.

\begin{table}
\centering
\small
\caption{A/B test results for web and app platforms}
\vspace{-4pt}
\begin{tabular}{@{}lcccc@{}}
\toprule
\textbf{Platform} & \textbf{CTR (\%)} & \textbf{OC (\%)} & \textbf{ATC/V (\%)} & \textbf{VP (\%)} \\ \midrule
\texttt{Web} & +4.9 & +2.8 & +1.5 & +2.2 \\ 
\texttt{App} & +2.4 & +0.23 & +1.10 & +0.2 \\ 
\bottomrule
\end{tabular}
\label{tab:commerce_metrics}
\end{table}

\section{Experiments}

Offline comparison across contextual bandit algorithms is inherently infeasible, as logged data only reflect rewards for actions taken by the deployed policy, leaving counterfactuals unobserved. Consequently, rigorous cross-policy evaluation necessitates live A/B experimentation, which we employ as the standard approach for model validation. We evaluated the proposed LinUCB-based dynamic carousel reordering in a live A/B test on a high-traffic product detail page. The test ran for almost two weeks, with 30\% traffic split evenly between control and treatment groups. The control condition used a static, rule-based carousel ordering fixed across categories, while the treatment condition employed dynamic ordering driven by the LinUCB policy. Each session was instrumented to capture displays, clicks, add-to-carts, and purchase events, along with engagement metrics such as bounce rate and demand per visitor (DPV). The primary performance indicator was Click Through Rate (\textbf{CTR}). Secondary metrics included Order Conversion (\textbf{OC}), Add-to-Carts per Visitor (\textbf{ATC/V}) \& the fraction of visitors completing purchases (\textbf{VP}). Daily averages were computed for both control and treatment variants, and performance lift was measured as the relative percentage improvement of treatment over control. The LinUCB-based variant consistently performed similar to or better than the heuristic baseline across all commerce metrics, with particularly strong gains in clicks, add to carts, and demand.

As shown in Table~\ref{tab:commerce_metrics}, the LinUCB-based variant outperformed the heuristic baseline on both platforms, with the reward function optimized primarily for CTR. The stronger lift on web (+4.9\% CTR, +2.8\% OC) compared to app reflects fundamental platform differences: web visitors arrive predominantly unlogged-in and often as new guests with minimal context, while app users are typically logged-in, loyal and with richer personalization signals. All reported numbers are statistically significant. 

\section{Discussion}
\begin{figure}[tb]
  \centering
  \includegraphics[width=\linewidth]{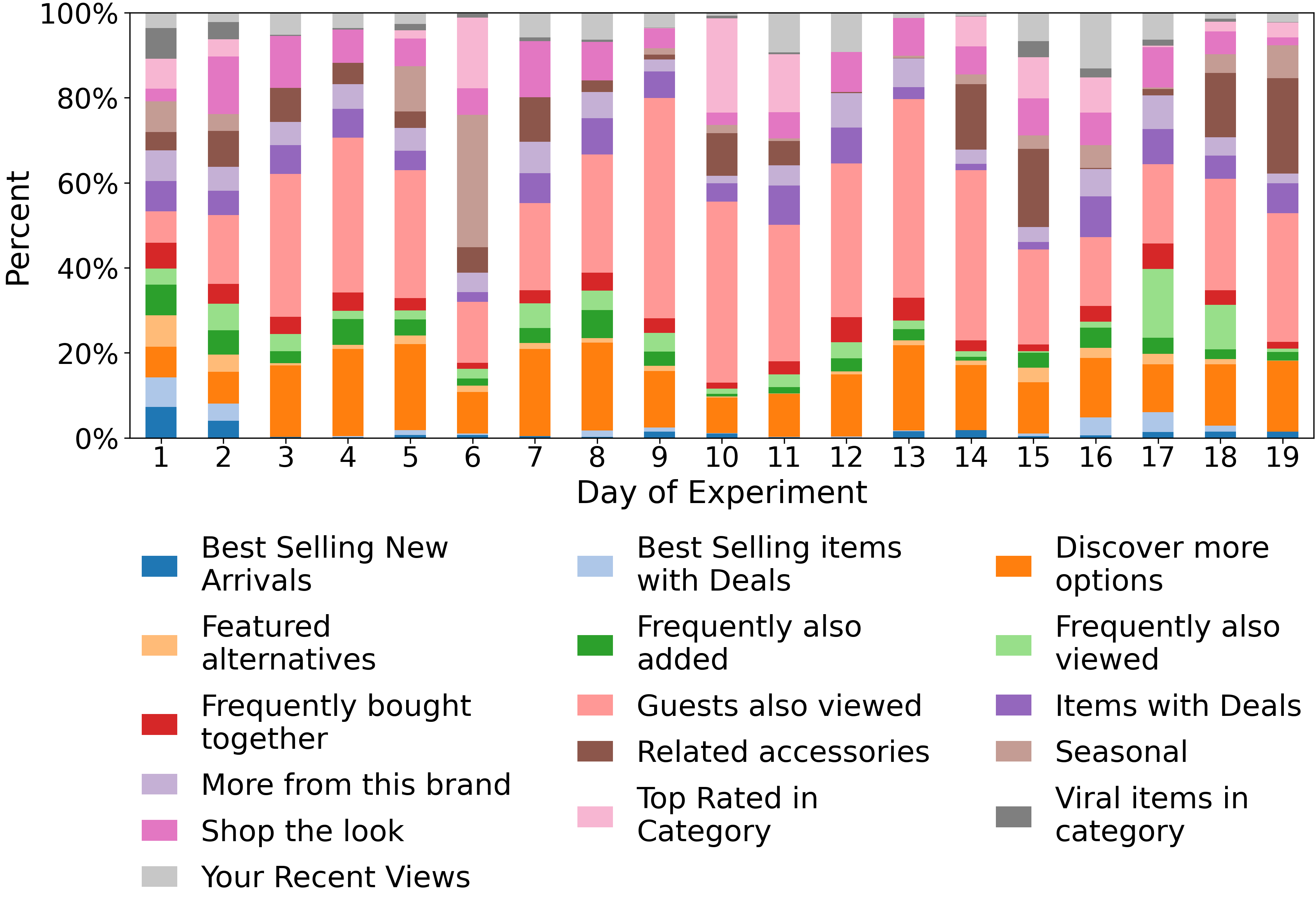}
  \vspace{-10pt} 
  \caption{Percent of time an arm was chosen as the top placement by day of experiment.}
  \label{fig:top_arms}
  \vspace{-11pt}
\end{figure}

The observed gains in \textbf{OC} and \textbf{DPV} suggest the policy effectively identified higher-converting layouts in real time, while the steady Bounce Rate across days confirms that the model did not harm user experience.
Performance trends indicate that the contextual bandit model improved conversion and engagement metrics for externally acquired traffic, traditionally characterized by low intent and high bounce behavior. Figure~\ref{fig:top_arms} illustrates how the distribution of top-ranked arms selected by the bandit evolved over the course of the experiment. After an initial exploration phase in the first few days, the policy shifted traffic to higher-performing carousels, disproportionately selecting "Users also Viewed" and "Discover More Options", while selecting "Best Selling New Arrivals" as the top arm less frequently. Note that selection is context-dependent and LinUCB quickly tailors which carousels are surfaced to each context. Rather than relying on a fixed placement, the bandit adapts dynamically, sometimes preferring "Seasonal" and "Related accessories" when the context warrants it. For example, when a user searches for "Apple Watch", strong brand-intent led the policy to surface a "Best-selling Apple" carousel prominently; for items currently experiencing a promotion, a "Save on similar items" carousel was displayed instead, emphasizing the value proposition of the sale. These shifts reflect a dynamic adjustment of the user experience based on the contextual signal.

\begin{figure}[t]
  \centering
  \hspace*{-0.1\columnwidth}
  \begin{minipage}[b]{0.5\columnwidth}
    \centering
    \begin{tikzpicture}
      \begin{axis}[
        width=1.08\linewidth,
        height=4.8cm,
        xlabel={Day},
        ylabel={Regret},
        xlabel style={font=\footnotesize},
        ylabel style={font=\footnotesize},
        xticklabel style={font=\footnotesize},
        yticklabel style={font=\footnotesize,/pgf/number format/fixed},
        axis x line*=bottom,
        axis y line*=left,
        ymin=0, ymax=0.28,
        xmin=0.5, xmax=9.5,
        xtick={1,2,3,4,5,6,7,8,9},
        grid=none
      ]
        \addplot+[mark=o,thick] coordinates {
          (1,0.270)(2,0.250)(3,0.255)(4,0.030)
          (5,0.025)(6,0.020)(7,0.025)(8,0.038)(9,0.035)
        };
      \end{axis}
    \end{tikzpicture}
    \vspace{-4pt}
    \textbf{(a)}
  \end{minipage}
  \hspace{-3.0mm}
  \begin{minipage}[b]{0.5\columnwidth}
    \centering
    \hspace*{0.02\columnwidth}%
    \begin{tikzpicture}
      \begin{axis}[
        width=1.08\linewidth,
        height=4.8cm,
        xlabel={Day},
        ylabel={Add-t0cart \% Lift},
        xlabel style={font=\footnotesize},
        ylabel style={font=\footnotesize},
        xticklabel style={font=\footnotesize},
        yticklabel style={font=\footnotesize,/pgf/number format/fixed},
        axis x line*=bottom,
        axis y line*=left,
        ymin=-6, ymax=12,
        xmin=0.5, xmax=12.5,
        xtick={1,2,3,4,5,6,7,8,9,10,11,12},
        grid=major, grid style={dashed,gray!30}
      ]
        \addplot+[mark=none,thick] coordinates {
          (1,-4.0)(2,3.0)(3,2.5)(4,2.5)(5,9.0)
          (6,1.5)(7,-0.5)(8,5.5)(9,4.0)(10,3.0)
          (11,2.5)(12,6.5)
        };
        \addplot[black,dashed,thin] coordinates {(0.5,0) (12.5,0)};
      \end{axis}
    \end{tikzpicture}
    \vspace{-4pt}
    \textbf{(b)}
  \end{minipage}

  \caption{(a) Daily avg. pseudo regret for bandit. (b) Daily \% lift in add-to-carts per visitor.}
  \label{fig:ab_plots}
\end{figure}
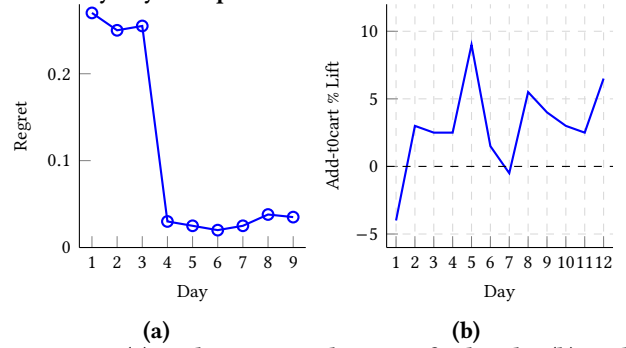

Figure ~\ref{fig:ab_plots} (a) displays the daily 
regret for the web test. It rises early during exploration and flattens out after 3 days, a 
pattern consistent with the policy identifying and exploiting high-performing orderings based on context. 
At interaction $t$, the agent observes an eligible set of arms $A_t$ and context vectors
$\{x_{t,a}\in\mathbb{R}^d : a\in A_t\}$, selects an arm $a_t\in A_t$, and receives a reward $r_t$.
Since counterfactual rewards for unchosen arms are not observed in logged bandit data,
we report a model-based pseudo-regret proxy computed from per-arm outcome models.
Let $\hat{\mu}_t(x_{t,a})$ denote the predicted expected reward of arm $a$ under a logistic regression
model trained on data strictly prior to time $t$ (forward chaining). We define the per-event proxy as $\hat{\Delta}_t = \max_{a \in A_t} \hat{\mu}_t(x_{t,a}) - \hat{\mu}_t(x_{t,a_t})$
and plot the daily average $\frac{1}{|\mathcal{T}_d|}\sum_{t\in\mathcal{T}_d}\hat{\Delta}_t$
(and cumulative $\sum_{t=1}^T \hat{\Delta}_t$). Figure \ref{fig:ab_plots} (b) illustrates \% lift in ATC/V, suggesting that 
LinUCB-driven variation consistently matched or outperformed the control across the experiment. 
The treatment maintained a modest but persistent 
edge, peaking at day 5—coinciding with increased 
site activity over the weekend. DPV \& CTR also 
followed a similar pattern, indicating that 
higher engagement translated into incremental 
commercial value rather than short-term exploration noise. Overall, these results suggest that contextual bandits offer a practical and scalable path toward adaptive page design, enabling e-commerce interfaces to continuously learn from user interactions and tailor layouts to user context.
We view this work as a first step toward a broader vision of adaptive, full-page layout optimization, where contextual bandits can continuously learn and personalize not only which modules are shown, but how the overall user experience is composed.

\section{Future Work}
While the proposed system demonstrates that contextual bandits can effectively optimize recommendation module layouts in a production environment, several opportunities remain for future improvement.

First, our current implementation employs Inverse Propensity Weighting (IPW) to partially correct for position bias. Although IPW is simple to implement and integrates naturally into a production pipeline, it is known to exhibit high variance when propensity scores are small. Future iterations of the system will investigate more robust off-policy estimators, such as 'Self-Normalized Inverse Propensity Scoring' and 'Doubly Robust' estimation, which have been shown to provide lower-variance and more reliable reward estimates.

Second, the current reward function is optimized primarily for click-through rate (CTR). While CTR serves as an effective proxy for user engagement, future work will explore multi-objective reward functions that directly optimize downstream business metrics such as Add-to-Cart, Order Conversion, and Demand Per Visitor. Such formulations would better align online learning with long-term business objectives.

Finally, the proposed architecture naturally supports continuous evolution of both the model and its contextual features. As richer behavioral signals become available, additional user, session, and product features can be incorporated without requiring changes to the serving infrastructure. Future work will ultimately investigate whole-page optimization, where multiple page components are jointly optimized rather than ranking recommendation modules independently.
 
 \textbf{Acknowledgments:} Authors would like to thank AJ Dabruzzi for coordinating with stakeholders \& Sujitha Chinta for visualizations.



\appendix

\end{document}